\documentclass[a4paper,fleqn]{cas-dc}

\usepackage[numbers]{natbib}
\usepackage[numbers]{natbib}

\usepackage{graphicx}
\usepackage{booktabs}
\usepackage{multirow}
\usepackage{caption}
\usepackage{subcaption}
\usepackage{placeins}

\def\tsc#1{\csdef{#1}{\textsc{\lowercase{#1}}\xspace}}
\tsc{WGM}
\tsc{QE}
\tsc{EP}
\tsc{PMS}
\tsc{BEC}
\tsc{DE}
\newcommand{\inlinefigure}[4][\linewidth]{%
\par\medskip
\noindent
\begin{minipage}{\linewidth}
\centering
\includegraphics[width=#1]{#2}
\captionof{figure}{#3}
\label{#4}
\end{minipage}
\par\medskip
}

\begin{document}

\shorttitle{Using Explainability as a Training-Time Reliability Signal for Efficient ECG Classification}
\shortauthors{V. K. Dangeti, S. N. Gowda}
%\begin{frontmatter}

\title [mode = title]{Using Explainability as a Training-Time Reliability Signal for Efficient ECG Classification}                      

%\tnotemark[1,2]

%\tnotetext[1]{This document is the results of the research project funded by the National Science Foundation.}

%\tnotetext[2]{The second title footnote which is a longer text matterto fill through the whole text width and overflow into another line in the footnotes area of the first page.}

\author[uon]{Veerendhra Kumar Dangeti}
\ead{psxvd1@nottingham.ac.uk}
\credit{Conceptualization of this study, Methodology, Software}

\author[oxford]{Xiao Gu}
\ead{xiao.gu@eng.ox.ac.uk}

\author[ningbo]{Ying Weng}
\ead{Ying.Weng@nottingham.edu.cn}

\author[uon]{Shreyank N Gowda}
\cormark[1]
\ead{shreyank.narayanagowda@nottingham.ac.uk}

\affiliation[uon]{
organization={School of Computer Science, University of Nottingham},
addressline={Jubilee Campus, Wollaton Road},
city={Nottingham},
postcode={NG8 1BB},
country={United Kingdom}
}

\affiliation[oxford]{
organization={Institute of Biomedical Engineering, Department of Engineering Science, University of Oxford},
city={Oxford},
postcode={OX3 7DQ},
country={United Kingdom}
}

\affiliation[ningbo]{
organization={School of Computer Science, University of Nottingham Ningbo China},
city={Ningbo},
postcode={315100},
country={China}
}

\cortext[cor1]{Corresponding author}

%\fntext[fn1]{This is the first author footnote, but is common to third author as well.}
%\fntext[fn2]{Another author footnote, this is a very long footnote and   it should be a really long footnote. But this footnote is not yet   sufficiently long enough to make two lines of footnote text.}

\begin{abstract}
Training deep neural networks for clinical time-series analysis is computationally demanding, yet many healthcare settings lack the resources required for repeated model development and deployment. This challenge is particularly evident in electrocardiogram classification, where large datasets and long training schedules make efficiency practically important. Progressive Data Dropout reduces training cost by excluding samples from gradient updates once they are learned, but it relies on model confidence and may retain samples that are difficult due to noise or ambiguity rather than useful signal. In this work, we introduce ERTS, an explainability-based reliability training signal for efficient ECG classification. ERTS uses explanation quality during training to distinguish between informative and unreliable uncertainty. Building on progressive data selection, we compute Grad-CAM attention maps for candidate samples and derive a focus score that measures whether model predictions are supported by coherent and localised patterns. Samples with low focus are filtered out, while those with meaningful attention are prioritised for gradient updates. We evaluate ERTS across three ECG datasets and multiple backbone architectures, showing consistent improvements in macro-F1 alongside reduced effective training cost. These results suggest that explanation quality can serve as a practical signal for improving both efficiency and reliability in clinical time-series learning. Code will be released.
\end{abstract}

\iffalse
\begin{graphicalabstract}
\includegraphics{figs/cas-grabs.pdf}
\end{graphicalabstract}

\begin{highlights}
\item Introduces ERTS, an explainability-based reliability training signal that uses explanation quality to guide sample selection during training.
\item Enhances training efficiency by filtering unreliable uncertain samples, reducing redundant gradient updates without modifying model architecture.
\item Demonstrates consistent improvements in macro-F1 and effective training cost across three ECG datasets and multiple backbone architectures.
\item Provides insight into model behaviour through class-distribution analysis, showing better retention of diagnostically relevant samples compared to confidence-only methods.
\item Establishes explainability as a training-time signal, extending its role beyond post hoc interpretation to improve both efficiency and reliability in clinical time-series learning.
\end{highlights}
\fi


\maketitle

\section{Introduction}

Deep learning has become a central tool in biomedical signal analysis, achieving strong performance on tasks such as electrocardiogram (ECG) classification, arrhythmia detection, and broader clinical prediction from physiological time-series data \cite{wagner2020ptb,strodthoff2020deep,cpsc2018}. Modern neural architectures can learn rich representations directly from multi-lead ECG recordings, enabling automated diagnostic support in real-world healthcare settings \cite{strodthoff2020deep}. However, these gains come with increasing computational cost. Training deep models typically requires long optimisation schedules, repeated experimentation, and substantial hardware resources, which are often limited in clinical environments. Tthe environmental and financial cost of training deep models has become an increasing concern, with recent work highlighting the carbon footprint and resource demands associated with modern machine learning pipelines \cite{strubell2019energy,schwartz2020greenai,patterson2021carbon,gowda2024watt}. This is particularly relevant in healthcare, where models may need to be retrained across datasets, institutions, and evolving diagnostic requirements. Several recent studies have demonstrated that deep neural networks can achieve cardiologist-level performance on arrhythmia detection and ECG interpretation tasks, highlighting their potential for large-scale clinical deployment \cite{hannun2019cardiologist,rajpurkar2017cardiologist,attia2019artificial,gu2026cardiac}.

Most efforts to address this challenge have focused on model-level efficiency, including lightweight architectures, pruning, quantisation, and knowledge distillation \cite{tan2019efficientnet,sandler2018mobilenetv2,han2015deep,jacob2018quantization,hinton2015distilling}. While effective for reducing model size and inference cost, these approaches do not directly address inefficiencies during training. Standard supervised learning exposes all samples to gradient updates across all epochs, implicitly assuming that each example remains equally informative throughout optimisation. In practice, many samples become easy early in training and contribute little new learning signal later, while a smaller subset continues to influence decision boundaries or reveal failure modes. Treating all samples uniformly can therefore lead to redundant computation.

This observation has motivated data-centric approaches that aim to prioritise informative samples or reduce unnecessary gradient updates. Methods such as curriculum learning, hard example mining, importance sampling, and dataset pruning explore different ways of adapting training to the varying utility of samples \cite{bengio2009curriculum,katharopoulos2018not,paul2021deep,yuan2025instancedependent,add}. Recent empirical studies further suggest that a substantial fraction of training data may contribute limited value once a model has already learned them, motivating approaches that dynamically identify and prioritise informative samples during optimisation \cite{toneva2018empirical,swayamdipta2020dataset,shriramprogressive,gowda2023synthetic}. More recently, Progressive Data Dropout (PDD) has shown that a simple training-time strategy can significantly reduce computation by progressively excluding samples from gradient updates once they are considered learned, while reintroducing the full dataset during a final revision stage \cite{shriramprogressive}. Despite its simplicity and effectiveness, PDD relies on model confidence as a proxy for usefulness. This issue is closely related to broader observations in noisy-label learning, where difficult examples may either reflect informative edge cases or arise from annotation errors and data ambiguity \cite{reed2014training,arazo2019unsupervised,northcutt2021confident}.

\begin{figure*}[!t]
\centering
\includegraphics[width=\linewidth]{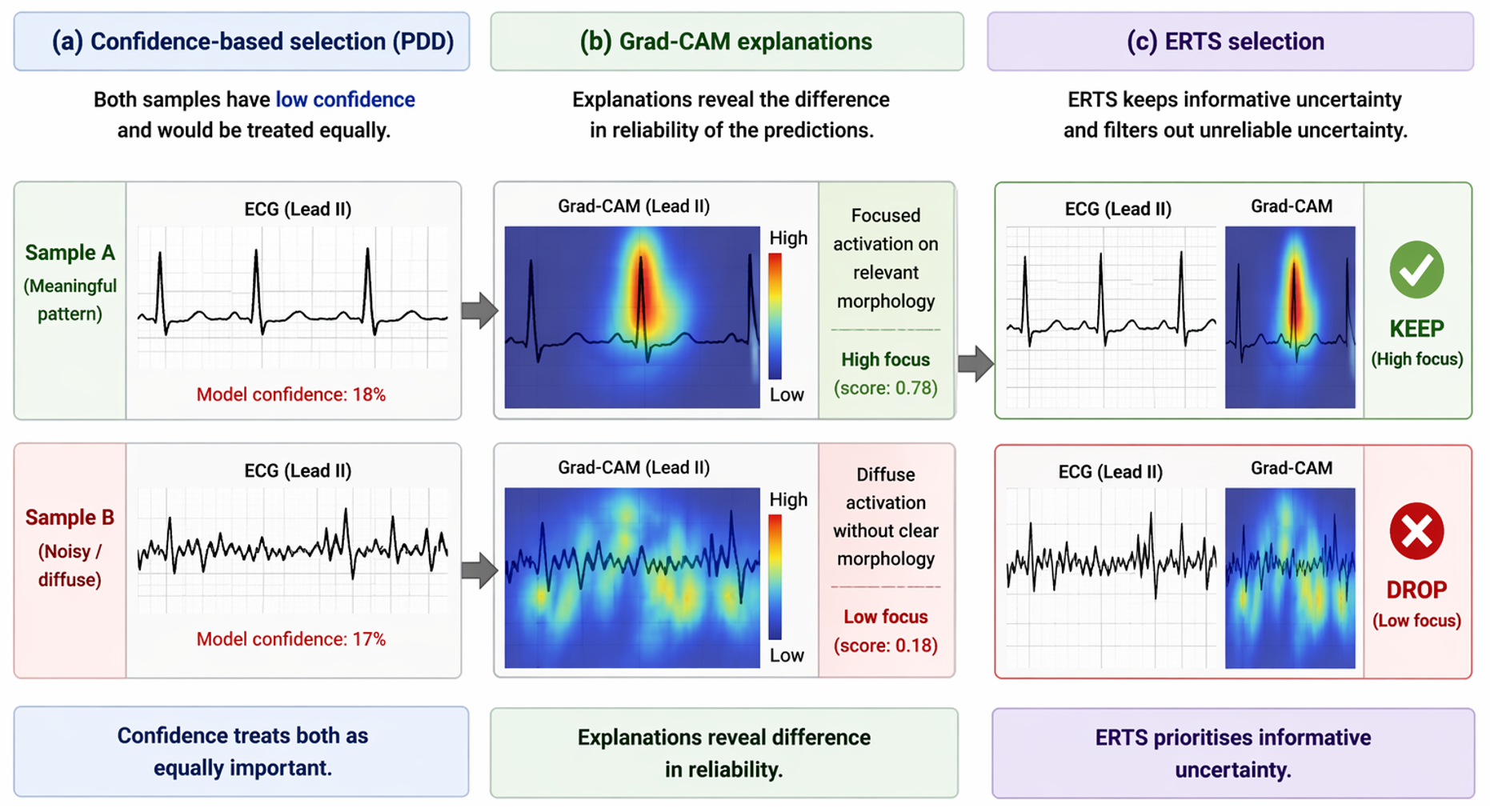}
\caption{
Standard confidence-based data selection treats all uncertain samples as equally informative. However, uncertainty may arise from meaningful but under-learned patterns or from noise and ambiguity. We use explanation quality to distinguish between these cases. Samples with focused and clinically meaningful attention are retained, while those with diffuse or unreliable explanations are filtered out, leading to more efficient and reliable training.
}
\label{fig:teaser}
\end{figure*}

In clinical data, however, uncertainty is not always informative. A sample may remain difficult because it contains subtle or underrepresented patterns that are important for learning, but it may also remain difficult because it is noisy, weakly labelled, or poorly aligned with clinically meaningful signal. Confidence-based selection cannot distinguish between these cases. As a result, it may continue to allocate computation to unreliable samples while discarding examples that are diagnostically informative but learned early. This distinction is illustrated in Fig.~\ref{fig:teaser}, where two samples with similar uncertainty exhibit fundamentally different explanatory characteristics, motivating the need for explanation-guided selection. This limitation is particularly important in ECG classification, where datasets often exhibit label ambiguity, inter-patient variability, and acquisition artefacts \cite{wagner2020ptb,strodthoff2020deep,wei2021learning}. In this setting, training efficiency is closely tied to reliability. 

Explainable artificial intelligence (xAI) offers a potential way to address this gap. Saliency-based methods such as Grad-CAM are widely used to identify which regions of an input influence a model's prediction \cite{selvaraju2017grad,chattopadhay2018grad}. Beyond Grad-CAM, a wide range of attribution methods have been proposed, including gradient-based, perturbation-based, and game-theoretic approaches, each offering different perspectives on how models utilise input features \cite{smilkov2017smoothgrad,sundararajan2017axiomatic,lundberg2017unified,montavon2018methods}. In ECG analysis, these methods have primarily been used post hoc to assess whether model attention aligns with clinically meaningful waveform morphology \cite{strodthoff2020deep,xai1,xai2,xai3}. However, this perspective suggests a broader opportunity. If explanation quality reflects whether model predictions are grounded in meaningful signal, then it may be possible to use explainability during training to guide which samples deserve further optimisation. More recently, there has been growing interest in using explanations not only for interpretation but also to guide model behaviour, for example through explanation regularisation or alignment with human-understandable evidence \cite{ross2017right,erion2021improving}.

In this work, we explore this idea through \textbf{ERTS}, an Explainability-based Reliability Training Signal for efficient ECG classification. ERTS augments progressive data selection with an explanation-guided filter that distinguishes between informative and unreliable uncertainty. Candidate samples are first identified using standard progressive data dropout criteria, after which Grad-CAM is used to assess whether the model's prediction is supported by coherent and localised evidence. A focus score derived from the explanation map is then used to filter out samples with diffuse or low-quality attention, while retaining those that are both uncertain and meaningfully grounded.

This work is guided by three research questions. Can explanation quality serve as a reliable signal for distinguishing informative from unreliable uncertainty during training. Does incorporating this signal improve the trade-off between efficiency and predictive performance across datasets and model architectures. Can explanation-guided selection provide interpretable insight into which samples contribute most to learning. To answer these questions, we evaluate ERTS across multiple ECG datasets and backbone architectures.

Our results show that ERTS consistently improves macro-F1 while reducing training cost compared to confidence-based selection alone. Analysis of retained samples further indicates that the method preferentially removes low-focus normal recordings while preserving diagnostically relevant hard cases. These findings suggest that explainability can play a practical role beyond post hoc interpretation, acting as a training-time signal that improves both efficiency and reliability in clinical time-series learning.

\section{Related Work}

\paragraph{Deep learning for ECG classification.}
Deep learning has become a dominant approach for automated ECG analysis, with convolutional and hybrid neural architectures achieving strong results on arrhythmia detection, diagnostic classification, and broader clinical prediction tasks \cite{wagner2020ptb,strodthoff2020deep,cpsc2018}. Public benchmarks such as PTB-XL have played an important role in standardising evaluation and exposing the challenges of multi-label classification, inter-patient variability, and clinically meaningful generalisation \cite{wagner2020ptb,strodthoff2020deep}. These developments have established ECG classification as both a practically important and methodologically demanding setting for studying efficient and trustworthy clinical learning.

\paragraph{Efficiency in medical AI and deep learning.}
A large body of work has sought to reduce the computational burden of deep learning through architectural design, model compression, and deployment-time optimisation. Representative directions include lightweight architectures such as EfficientNet and MobileNet, as well as pruning, quantisation, and knowledge distillation \cite{tan2019efficientnet,sandler2018mobilenetv2,han2015deep,jacob2018quantization,hinton2015distilling}. In medical AI, these ideas are especially relevant because many healthcare settings lack access to large-scale computational infrastructure, making efficient model development and retraining an important practical concern. However, most of this work focuses on model efficiency rather than training-data efficiency, leaving open the question of how to reduce unnecessary computation during optimisation itself.

\paragraph{Data selection and training-time efficiency.}
Beyond model-centric efficiency, a growing line of work explores whether all training samples need to contribute equally throughout learning. Curriculum learning, hard example mining, importance sampling, and dataset pruning all share the intuition that training can be improved by prioritising informative examples or reducing redundancy \cite{bengio2009curriculum,shrivastava2016training,katharopoulos2018not,paul2021deep,yuan2025instancedependent}. Related methods such as Data Diet and other data pruning strategies show that substantial portions of training data may contribute limited value once the model has already learned them \cite{paul2021deep}. Progressive Data Dropout (PDD) extends this idea through a simple training-time policy that progressively excludes subsets of samples from gradient updates and restores the full dataset during a final revision stage \cite{shriramprogressive}. While such approaches can improve training efficiency, most rely on confidence, difficulty, or optimisation-based heuristics that do not explicitly account for whether hard samples are clinically meaningful or merely noisy.

\paragraph{Data selection under noisy or ambiguous labels.}
The distinction between informative and misleading hard examples is particularly important in settings with label noise or ambiguity. Prior work has shown that hard samples can be beneficial for learning, but may also reflect annotation error, domain shift, or low-quality data \cite{wang2018data,wei2021learning,yang2025dynamic}. This tension is especially relevant in healthcare, where ambiguity in labels, inter-rater disagreement, and acquisition artefacts are common \cite{noise1,noise2,noise3}. As a result, a sample that remains difficult throughout training is not necessarily valuable. This motivates selection criteria that go beyond confidence alone and ask whether model uncertainty is supported by meaningful signal.

\paragraph{Explainability in medical machine learning.}
Explainable artificial intelligence (xAI) has become an important tool for inspecting model behaviour in safety-critical domains. In medical imaging and signal analysis, saliency-based approaches such as Grad-CAM and Grad-CAM++ are commonly used to identify the input regions most responsible for a prediction \cite{selvaraju2017grad,chattopadhay2018grad}. In ECG analysis, explainability has largely been used in a post hoc manner to assess whether learned attention aligns with clinically plausible morphology or known diagnostic markers \cite{strodthoff2020deep,sobahi2022attention,xai1,xai2,xai3}. These methods are typically used for interpretation, trust assessment, or qualitative validation after training, rather than as part of the optimisation process itself.

\paragraph{Explainability as a training signal.}
A smaller but growing body of work suggests that explanations can be used not only to interpret predictions, but also to guide learning. In broader machine learning, attribution-based priors and explanation regularisation have been explored as ways to align model behaviour with desired evidence patterns \cite{erion2021improving,schramowski2020making}. However, this idea remains underexplored in clinical time-series learning, where explanation quality could provide a useful signal for distinguishing informative uncertainty from noise-driven uncertainty. Our work builds on this intuition by using explanation quality during training as a reliability signal for data selection. Rather than treating explainability as a purely post hoc tool, we use it to refine which uncertain samples continue to receive gradient updates, with the goal of improving both efficiency and trustworthiness in ECG learning.

\section{Method}

We study training-time data selection for efficient clinical time-series learning. We first introduce preliminaries and briefly recap Progressive Data Dropout (PDD). We then present \textbf{ERTS}, an explainability-based reliability training signal that refines sample selection using explanation quality.

\subsection{Preliminaries}

Let $\mathcal{D} = \{(x_i, y_i)\}_{i=1}^N$ denote a training dataset, where $x_i$ is an input (e.g., a 12-lead ECG signal) and $y_i$ is the corresponding label. Let $f_\theta$ be a neural network with parameters $\theta$. Standard supervised learning minimises:
\begin{equation}
\mathcal{L} = \frac{1}{N} \sum_{i=1}^{N} \ell(f_\theta(x_i), y_i).
\end{equation}

In this setting, all samples contribute to backpropagation at every epoch, implicitly assuming that each example remains equally informative throughout training. In practice, many samples become easy early and provide limited learning signal later, leading to redundant computation.

\subsection{Progressive Data Dropout (PDD)}

Progressive Data Dropout (PDD) addresses this inefficiency by reducing the number of samples used for gradient updates over time. At each epoch $t$, a subset $\mathcal{D}_t \subseteq \mathcal{D}$ is selected for backpropagation, while the full dataset can still be used for forward passes.

We follow the formulation in \cite{shriramprogressive}. A common variant (DBPD) uses model confidence. Let $p_t(x_i)$ denote the predicted probability for the correct class. A sample is considered easy if $p_t(x_i) \geq \tau$, where $\tau$ is a threshold. The retained subset is:
\begin{equation}
\mathcal{D}_t = \{(x_i, y_i) \in \mathcal{D} \mid p_t(x_i) < \tau\}.
\end{equation}

This progressively reduces the number of samples used for optimisation, improving efficiency. A final revision stage typically reintroduces the full dataset to maintain global coverage. However, in clinical data, confidence is not always a reliable proxy for usefulness. Samples may remain difficult either due to informative patterns or due to noise and ambiguity, making it challenging to distinguish which samples deserve continued optimisation.

\subsection{Explainability-Guided Data Selection (ERTS)}

\begin{figure*}[!t]
\centering
\includegraphics[width=\linewidth]{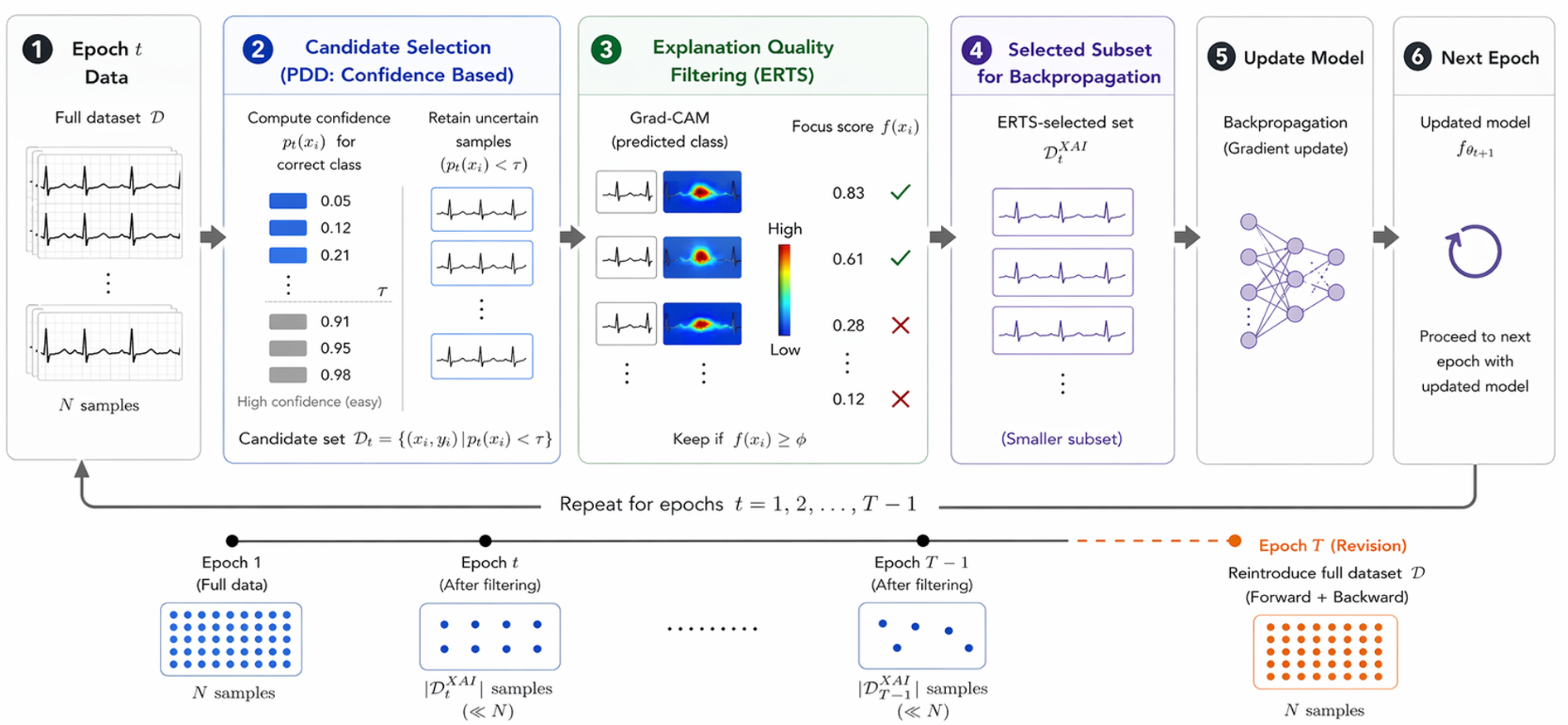}
\caption{
Overview of the proposed ERTS framework. At each epoch, candidate samples are first selected using confidence-based filtering (PDD). Explanation quality is then used to refine this subset by retaining samples with focused and reliable attention. The selected subset is used for backpropagation, and the model is updated iteratively. Over training, the effective dataset size decreases, and in the final epoch the full dataset is reintroduced for a revision step.
}
\label{fig:method}
\end{figure*}

To address this limitation, we propose our work \textbf{ERTS} (Explainability-based Reliability Training Signal), which uses explanation quality as a training-time reliability signal to refine data selection. An overview of the framework is shown in Fig.~\ref{fig:method}.

The key idea is to distinguish between two types of uncertainty: (i) informative uncertainty, where the model attends to meaningful structure but has not yet learned it, and (ii) unreliable uncertainty, where the model's attention is diffuse or poorly aligned with relevant signal.

\paragraph{Grad-CAM explanations.}
For each input $x_i$, we compute a Grad-CAM map $A(x_i)$ corresponding to the predicted class \cite{selvaraju2017grad}. In ECG data, this highlights temporal regions and lead interactions that contribute most to the model's prediction.

\paragraph{Focus score.}
We quantify explanation quality using a focus score that measures how concentrated the explanation is. Let $A(x_i)$ be normalised to $[0,1]$. We define:
\begin{equation}
f(x_i) = \frac{1}{|\Omega_i|} \sum_{w \in \Omega_i} A(x_i)[w],
\end{equation}
\begin{equation}
\Omega_i = \{w \mid A(x_i)[w] \geq Q_{90}(A(x_i))\},
\end{equation}
where $Q_{90}$ denotes the 90th percentile. This formulation emphasises the most salient regions of the explanation map, providing a measure of how concentrated the model's attention is. High values indicate focused, localised attention, while low values indicate diffuse or unstable attention.

\paragraph{Selection strategy.}
At each epoch, we first apply the PDD criterion to identify uncertain samples. We then refine this subset using the focus score:
\begin{equation}
\mathcal{D}_t^{\text{XAI}} = \{(x_i, y_i) \in \mathcal{D}_t \mid f(x_i) \geq \phi\},
\end{equation}
where $\phi$ is a threshold.

This two-stage selection separates informative uncertainty (low confidence, high focus) from unreliable uncertainty (low confidence, low focus). Only the former are prioritised for gradient updates.

\paragraph{Computational considerations.}
Grad-CAM is computed only for candidate samples selected by the PDD stage, which limits the additional computational overhead. In practice, this overhead is small relative to the savings obtained from reduced backpropagation.

\subsection{Training Procedure}

Training proceeds with dynamic sample selection at each epoch. For epochs $t = 1, \dots, T-1$, the model first computes predictions and confidence scores for all samples. A subset of uncertain samples is selected using PDD, followed by explainability-based filtering using the focus score. The resulting subset $\mathcal{D}_t^{\text{XAI}}$ is used for backpropagation to update model parameters.

In the final epoch $T$, the full dataset $\mathcal{D}$ is reintroduced for a revision step, ensuring that all samples contribute to the final model.

\subsection{Effective Epochs}

To quantify training efficiency independently of hardware, we use effective epochs (EE), defined as:
\begin{equation}
\text{EE} = \frac{\text{Total samples used for backpropagation}}{N}.
\end{equation}
This metric reflects the equivalent number of full passes over the dataset in terms of gradient updates and enables fair comparison across methods.

\section{Experimental Analysis}

We evaluate ERTS through a comprehensive set of experiments spanning three publicly used ECG datasets, PTB-XL, CPSC 2018, and Georgia 2020, and three backbone architectures with different capacity and efficiency profiles, namely EfficientNetV2-S, ResNet-18, and MobileNetV2. This setup allows us to assess whether the proposed explanation-guided filtering strategy generalises across datasets with different label characteristics, class distributions, and model families. For each setting, we compare the standard Progressive Data Dropout variants, including DBPD, SMRD, and SRD, against their ERTS-enhanced counterparts. Performance is evaluated using macro-F1 as the primary metric, given the class imbalance present in ECG classification, along with overall accuracy and Effective Epochs (EE) to quantify the trade-off between predictive performance and training efficiency.

\subsection{Overall Performance}

Table~\ref{tab:main_results} provides a consolidated comparison of the strongest configuration obtained for each dataset-backbone pair. For every dataset and architecture, we evaluate the family of standard PDD-based strategies, including difficulty-based progressive dropout (DBPD), schedule-matched random dropout (SMRD), and scalar random dropout (SRD). We then compare these against the corresponding ERTS-enhanced variants, where the same underlying data selection strategy is augmented with explanation-based reliability filtering. Thus, the table is not intended to report a single fixed variant across all settings, but rather to identify the best-performing member of each method family under each experimental condition.

The purpose of this comparison is to assess whether incorporating explanation quality improves the best achievable efficiency-performance trade-off, rather than only improving one particular dropout rule. This is important because DBPD, SMRD, and SRD differ in how they construct the training subset. DBPD uses confidence to retain difficult samples, SMRD follows a matched retention schedule with random selection, and SRD applies a scalar random decay. If ERTS only improved one of these cases, its usefulness would be more limited. Instead, Table~\ref{tab:main_results} asks a broader question: after allowing each method family to select its strongest variant, does explanation-guided filtering still provide an advantage?

Across all nine dataset-backbone combinations, the answer is consistently yes. The best ERTS variant achieves a higher macro-F1 score than the best plain PDD baseline in every case, while also requiring fewer effective epochs. This means that the improvement is not an artefact of a single dataset, architecture, or selection rule. The gains hold for larger backbones such as EfficientNetV2-S, standard architectures such as ResNet-18, and lightweight models such as MobileNetV2. They also hold across datasets with different label characteristics and class distributions. This consistency suggests that explanation quality provides a useful additional signal for deciding which uncertain samples should continue to contribute to gradient updates.

The table also highlights that ERTS improves both sides of the trade-off simultaneously. In each row, the ERTS-enhanced method increases macro-F1 while reducing EE, indicating that the method is not simply obtaining better performance by using more computation. Instead, it achieves better predictive performance with fewer effective passes through the training data. This supports the central claim of the paper: explanation quality can act as a training-time reliability signal that helps remove less useful uncertain samples while preserving those that remain informative for learning.

\begin{table*}[!t]
\centering
\caption{
Summary of the best standard PDD and ERTS-enhanced results across three ECG datasets and three backbone architectures. Each result is reported as macro-F1 / Effective Epochs (EE). $\Delta$F1 denotes the absolute macro-F1 improvement of ERTS over the best corresponding plain PDD configuration, while EE saved reports the relative reduction in effective epochs. Bold values indicate the best result for each dataset--backbone pair.
}
\label{tab:main_results}
\begin{tabular}{l l l l c c}
\toprule
Dataset & Backbone & Best PDD & Best ERTS & $\Delta$F1 & EE saved \\
\midrule
PTB-XL & EfficientNetV2-S & 0.6488 / 21.8 & \textbf{0.6586 / 18.9} & +0.0098 & 13.3\% \\
PTB-XL & ResNet-18        & 0.6539 / 22.3 & \textbf{0.6624 / 19.3} & +0.0085 & 13.5\% \\
PTB-XL & MobileNetV2      & 0.6336 / 21.6 & \textbf{0.6420 / 18.7} & +0.0084 & 13.4\% \\
\midrule
CPSC 2018 & EfficientNetV2-S & 0.7166 / 23.8 & \textbf{0.7188 / 19.4} & +0.0022 & 18.3\% \\
CPSC 2018 & ResNet-18        & 0.7052 / 23.8 & \textbf{0.7088 / 19.4} & +0.0036 & 18.5\% \\
CPSC 2018 & MobileNetV2      & 0.6833 / 23.1 & \textbf{0.6868 / 18.8} & +0.0035 & 18.6\% \\
\midrule
Georgia 2020 & EfficientNetV2-S & 0.6657 / 40.3 & \textbf{0.6736 / 38.2} & +0.0079 & 5.2\% \\
Georgia 2020 & ResNet-18        & 0.6521 / 40.3 & \textbf{0.6569 / 34.0} & +0.0048 & 15.7\% \\
Georgia 2020 & MobileNetV2      & 0.6319 / 39.1 & \textbf{0.6365 / 33.0} & +0.0046 & 15.6\% \\
\bottomrule
\end{tabular}
\end{table*}

Figure~\ref{fig:pareto} provides a complementary view of the efficiency--performance trade-off. Across all method families, ERTS variants move the Pareto frontier toward higher accuracy and lower effective epochs. In particular, XAI-DBPD and XAI-SMRD achieve improvements in predictive performance while reducing the effective number of training epochs by more than 65\% relative to the baseline. This indicates that explanation-guided filtering improves not only absolute performance but also the overall efficiency of optimisation.

\inlinefigure[0.95\linewidth]
{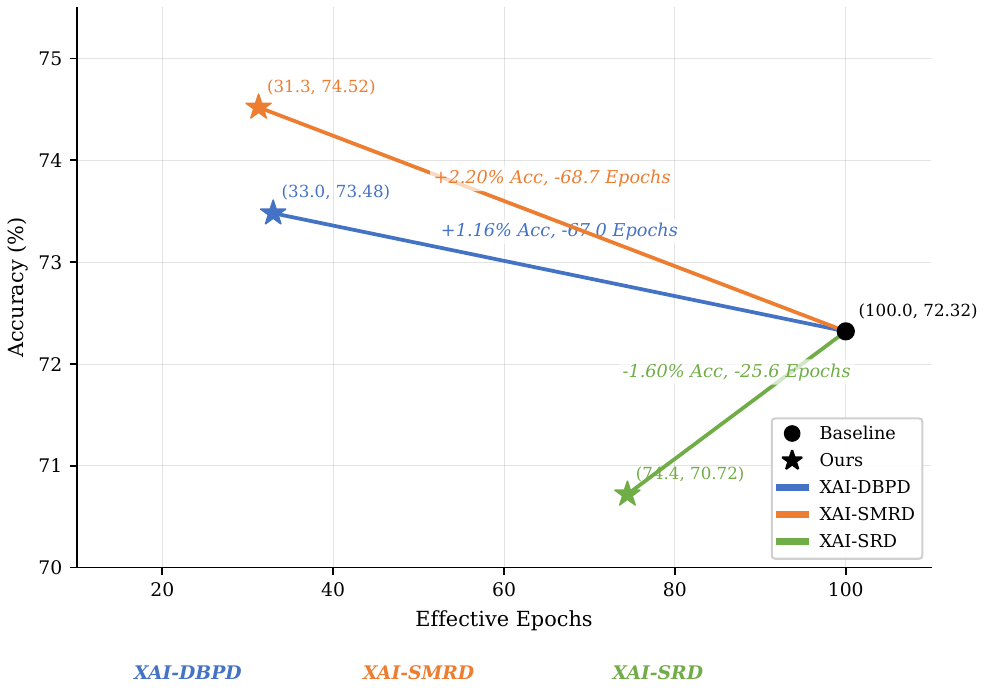}
{
Pareto comparison between predictive performance and training efficiency. Each point corresponds to the best-performing configuration within a method family. ERTS consistently shifts the trade-off frontier toward higher accuracy and lower effective epochs, demonstrating simultaneous gains in performance and training efficiency.
}
{fig:pareto}

\subsection{PTB-XL Results}

\begin{table*}[!t]
\centering
\caption{
Detailed PTB-XL results across EfficientNetV2-S, ResNet-18, and MobileNetV2. The table reports macro-F1, test accuracy, total samples used for backpropagation, Effective Epochs (EE), and the number of samples removed by the Grad-CAM explanation filter where applicable. We include the full-data baseline, standard PDD variants, and the corresponding ERTS-enhanced variants. Sample counts are abbreviated using K for thousands and M for millions. Missing or unavailable entries are denoted by ``--''. Bold values indicate the best macro-F1 within each backbone.
}
\label{tab:ptbxl_all_backbones_full}
\resizebox{\textwidth}{!}{%
\begin{tabular}{l
ccccc
ccccc
ccccc}
\toprule
\multirow{2}{*}{Model}
& \multicolumn{5}{c}{EfficientNetV2-S}
& \multicolumn{5}{c}{ResNet-18}
& \multicolumn{5}{c}{MobileNetV2} \\
\cmidrule(lr){2-6}
\cmidrule(lr){7-11}
\cmidrule(lr){12-16}
& F1 & Acc. & Samples & EE & CAM
& F1 & Acc. & Samples & EE & CAM
& F1 & Acc. & Samples & EE & CAM \\
\midrule

Baseline
& 0.6384 & 73.52 & 1.74M & 100.0 & --
& 0.6272 & 72.52 & 1.54M & 100.0 & --
& 0.6078 & 70.92 & 1.54M & 100.0 & -- \\

\midrule

DBPD ($\tau=0.3$)
& 0.4990 & 63.01 & 281K & 16.1 & --
& 0.5128 & 64.31 & 291K & 16.8 & --
& 0.4969 & 62.91 & 282K & 16.3 & -- \\

ERTS-DBPD ($\tau=0.3,\phi=0.7$)
& 0.5181 & 60.83 & 259K & 14.9 & 102K
& 0.5315 & 62.12 & 270K & 15.5 & 106K
& 0.5151 & 60.75 & 262K & 15.0 & 103K \\

DBPD ($\tau=0.7$)
& 0.6488 & \textbf{75.66} & 380K & 21.8 & --
& 0.6539 & 75.13 & 388K & 22.3 & --
& 0.6336 & 73.48 & 377K & 21.6 & -- \\

ERTS-DBPD ($\tau=0.7,\phi=0.5$)
& 0.5915 & 68.79 & 310K & 17.8 & 240K
& 0.6028 & 69.56 & 319K & 18.3 & 245K
& 0.5843 & 68.03 & 309K & 17.7 & 238K \\

ERTS-DBPD ($\tau=0.7,\phi=0.7$)
& 0.6418 & 74.12 & 324K & 20.6 & 244K
& 0.6462 & 74.58 & 332K & 21.0 & 248K
& 0.6262 & 72.94 & 322K & 20.4 & 241K \\

ERTS-DBPD ($\tau=0.7,\phi=0.9$)
& \textbf{0.6586} & 74.82 & 329K & \textbf{18.9} & 241K
& \textbf{0.6624} & 74.98 & 337K & \textbf{19.3} & 247K
& \textbf{0.6420} & 73.34 & 327K & \textbf{18.7} & 239K \\

\midrule

SMRD ($\tau=0.3$)
& 0.5674 & 74.16 & 327K & 18.8 & --
& 0.5782 & 74.83 & 336K & 19.2 & --
& 0.5603 & 73.18 & 325K & 18.6 & -- \\

ERTS-SMRD ($\tau=0.3,\phi=0.5$)
& 0.5396& 70.61& 291K& 16.7& 110k& 0.5496 & 71.23 & 298K & 17.1 & 112K
& 0.5325 & 69.66 & 290K & 16.6 & 109K \\

SMRD ($\tau=0.7$)
& 0.6548 & 74.70 & 477K & 27.4 & --& 0.6597 & \textbf{75.89} & 489K & 28.1 & --
& 0.6393 & \textbf{74.22} & 475K & 27.3 & -- \\

ERTS-SMRD ($\tau=0.7,\phi=0.7$)
& 0.6513& 73.71& 451K& 25.9& 290k& 0.6559 & 74.90 & 462K & 26.5 & 295K
& 0.6356 & 73.26 & 448K & 25.7 & 286K \\

\midrule

SRD ($\gamma=0.99$)
& 0.6405 & 75.43 & 1.09M & 62.4 & --
& 0.6458 & 75.73 & 1.10M & 64.8 & --
& 0.6259 & 74.06 & 1.07M & 62.9 & -- \\

ERTS-SRD
& 0.6479& 75.23& 1.24M& 59.6& 556k& 0.6549 & 75.49 & 1.04M & 59.1 & 564K
& 0.6347 & 73.83 & 1.01M & 57.3 & 547K \\

\bottomrule
\end{tabular}%
}
\end{table*}

On PTB-XL, ERTS provides a clear and detailed example of how explanation quality can improve training-time data selection. Table~\ref{tab:ptbxl_all_backbones_full} reports the results across three backbone architectures, namely EfficientNetV2-S, ResNet-18, and MobileNetV2. Unlike the summary comparison in Table~\ref{tab:main_results}, this table provides a more fine-grained view of the behaviour of individual PDD variants and their ERTS-enhanced counterparts. It includes the full-data baseline, DBPD, SMRD, and SRD, together with the corresponding explanation-guided variants where available. This allows us to examine not only whether ERTS improves the best-performing setting, but also how it affects different forms of progressive sample selection.

Across all three backbones, the strongest DBPD-based result is obtained by ERTS-DBPD with $\tau=0.7$ and $\phi=0.9$. For EfficientNetV2-S, standard DBPD improves over the full-data baseline, increasing macro-F1 from 0.6384 to 0.6488 while reducing the effective epoch count from 100.0 to 21.8. When explanation-based filtering is added, ERTS-DBPD further improves macro-F1 to 0.6586 and reduces EE to 18.9. A similar pattern is observed for ResNet-18, where DBPD achieves 0.6539 macro-F1 at 22.3 EE, while ERTS-DBPD reaches 0.6624 macro-F1 at 19.3 EE. MobileNetV2 follows the same trend, with DBPD obtaining 0.6336 macro-F1 at 21.6 EE and ERTS-DBPD improving this to 0.6420 macro-F1 at 18.7 EE. These results show that the benefit of ERTS is not specific to a single architecture, but is consistent across larger, standard, and lightweight backbones.

The table also highlights the importance of the explanation-filtering threshold. The most aggressive setting, $\phi=0.5$, substantially reduces the number of effective epochs but consistently harms predictive performance. For example, under EfficientNetV2-S, ERTS-DBPD with $\phi=0.5$ reduces EE to 17.8 but decreases macro-F1 to 0.5915. The same degradation appears for ResNet-18 and MobileNetV2, where macro-F1 falls to 0.6028 and 0.5843, respectively. This suggests that overly aggressive explanation filtering removes not only unreliable samples but also informative hard cases that remain valuable for learning. In contrast, $\phi=0.7$ gives a more moderate trade-off, but still does not consistently outperform plain DBPD. The best setting is $\phi=0.9$, which acts as a more selective reliability filter. It removes low-focus samples while preserving uncertain examples that still contain useful explanatory structure.

The SMRD and SRD results provide additional context. For SMRD, the plain variant remains competitive in macro-F1, especially for ResNet-18 and MobileNetV2. However, the corresponding ERTS-SMRD variants reduce the total number of samples and effective epochs, indicating that explanation-guided filtering can still improve training efficiency even when the performance gain is smaller. Similarly, ERTS-SRD reduces the effective epoch count compared with SRD and improves macro-F1 for the available ResNet-18 and MobileNetV2 results. This is important because SRD is not confidence-driven in the same way as DBPD. The fact that explanation filtering remains useful in this setting suggests that the Grad-CAM focus signal captures information beyond model confidence alone.

Overall, Table~\ref{tab:ptbxl_all_backbones_full} supports three observations. First, progressive data selection already provides substantial efficiency gains over full-data training. Second, adding explanation quality as a reliability signal further improves the efficiency--performance trade-off, most clearly for DBPD. Third, the strength of the explanation filter must be chosen carefully. ERTS works best not as an aggressive pruning mechanism, but as a selective filter that removes samples with weak or diffuse explanatory evidence while retaining informative uncertainty. This aligns with the central motivation of the method: the goal is not simply to train on fewer samples, but to identify which uncertain samples are useful for learning and which are likely to reflect noise, ambiguity, or poor explanatory structure.

\subsection{CPSC 2018 Results}

\begin{table*}[!t]
\centering
\caption{
Detailed CPSC 2018 results across EfficientNetV2-S, ResNet-18, and MobileNetV2. The table reports macro-F1, test accuracy, total samples used for backpropagation, Effective Epochs (EE), and the number of samples removed by the Grad-CAM explanation filter where applicable. We include the full-data baseline, standard PDD variants, and their corresponding ERTS-enhanced variants. Sample counts are abbreviated using K for thousands. Missing or unavailable entries are denoted by ``--''. Bold values indicate the best macro-F1 within each backbone.
}
\label{tab:cpsc_all_backbones_full}
\resizebox{\textwidth}{!}{%
\begin{tabular}{l
ccccc
ccccc
ccccc}
\toprule
\multirow{2}{*}{Model}
& \multicolumn{5}{c}{EfficientNetV2-S}
& \multicolumn{5}{c}{ResNet-18}
& \multicolumn{5}{c}{MobileNetV2} \\
\cmidrule(lr){2-6}
\cmidrule(lr){7-11}
\cmidrule(lr){12-16}
& F1 & Acc. & Samples & EE & CAM
& F1 & Acc. & Samples & EE & CAM
& F1 & Acc. & Samples & EE & CAM \\
\midrule

Baseline
& 0.6720 & 71.51 & 550K & 100.0 & --
& 0.6610 & 70.70 & 550K & 100.0 & --
& 0.6405 & 69.14 & 550K & 100.0 & -- \\

\midrule

DBPD ($\tau=0.3$)
& 0.6693 & 71.22 & 104K & 18.81 & --
& 0.6578 & 70.41 & 104K & 18.81 & --
& 0.6374 & 68.86 & 101K & 18.32 & -- \\

ERTS-DBPD ($\tau=0.3,\phi=0.7$)
& 0.6670 & 71.10 & 82K & 14.80 & 22K
& 0.6605 & 70.60 & 81K & 14.70 & 23K
& 0.6412 & 69.05 & 78K & 14.28 & 22K \\

DBPD ($\tau=0.7$)
& 0.6813 & 71.87 & 137K & 24.82 & --
& 0.6708 & 71.29 & 137K & 24.82 & --
& 0.6500 & 69.72 & 132K & 24.13 & -- \\

ERTS-DBPD ($\tau=0.7,\phi=0.5$)
& 0.6752 & 71.48 & 97K & 17.67 & 39K
& 0.6654 & 70.93 & 96K & 17.49 & 40K
& 0.6448 & 69.37 & 93K & 16.98 & 39K \\

ERTS-DBPD ($\tau=0.7,\phi=0.7$)
& 0.6834 & 72.01 & 111K & 20.25 & 25K
& 0.6735 & 71.48 & 112K & 20.29 & 25K
& 0.6526 & 69.90 & 108K & 19.68 & 24K \\

ERTS-DBPD ($\tau=0.7,\phi=0.9$)
& 0.6861 & 72.19 & 129K & 23.38 & 8K
& 0.6764 & 71.62 & 129K & 23.43 & 8K
& 0.6554 & 70.04 & 125K & 22.74 & 7K \\

\midrule

SMRD ($\tau=0.3$)
& 0.7166 & 74.71 & 131K & 23.76 & --
& 0.7052 & 73.98 & 131K & 23.76 & --
& 0.6833 & 72.37 & 127K & 23.07 & -- \\

ERTS-SMRD ($\tau=0.3,\phi=0.7$)
& \textbf{0.7188} & \textbf{74.82} & 107K & \textbf{19.43} & 24K
& \textbf{0.7088} & \textbf{74.20} & 107K & \textbf{19.39} & 24K
& \textbf{0.6868} & \textbf{72.58} & 104K & \textbf{18.82} & 24K \\

SMRD ($\tau=0.7$)
& 0.7099 & 74.27 & 171K & 31.17 & --
& 0.6996 & 73.61 & 171K & 31.17 & --
& 0.6779 & 71.99 & 166K & 30.23 & -- \\

ERTS-SMRD ($\tau=0.7,\phi=0.7$)
& 0.7132 & 74.40 & 137K & 24.87 & 35K
& 0.7037 & 73.88 & 137K & 24.90 & 35K
& 0.6819 & 72.27 & 133K & 24.21 & 33K \\

\midrule

SRD ($\gamma=0.95$)
& 0.6972 & 74.13 & 111K & 20.11 & --
& -- & -- & -- & -- & --
& -- & -- & -- & -- & -- \\

ERTS-SRD ($\gamma=0.95,\phi=0.7$)
& 0.7001 & 74.20 & 91K & 16.51 & 20K
& -- & -- & -- & -- & --
& -- & -- & -- & -- & -- \\

SRD ($\gamma=0.98$)
& 0.6978 & 74.32 & 111K & 20.32 & --
& 0.6896 & 73.54 & 111K & 20.32 & --
& 0.6682 & 71.91 & 107K & 19.72 & -- \\

ERTS-SRD ($\gamma=0.98,\phi=0.7$)
& 0.7010 & 74.41 & 91K & 16.60 & 19K
& 0.6931 & 73.71 & 91K & 16.58 & 20K
& 0.6716 & 72.08 & 88K & 16.09 & 19K \\

SRD ($\gamma=0.99$)
& 0.7035 & 72.97 & 544K & 98.89 & --
& 0.6948 & 72.21 & 544K & 21.13 & --
& 0.6732 & 70.62 & 528K & 20.50 & -- \\

ERTS-SRD ($\gamma=0.99,\phi=0.7$)
& 0.7062 & 73.12 & 424K & 77.16 & 120K
& 0.6972 & 72.43 & 447K & 17.39 & 96K
& 0.6756 & 70.83 & 434K & 16.87 & 93K \\

\bottomrule
\end{tabular}%
}
\end{table*}

Table~\ref{tab:cpsc_all_backbones_full} reports the full CPSC 2018 results across EfficientNetV2-S, ResNet-18, and MobileNetV2. As with the PTB-XL analysis, we include the baseline, standard PDD variants, and the corresponding ERTS-enhanced configurations wherever available. This allows us to examine whether explanation-guided filtering remains beneficial on a dataset with different label characteristics and overall performance behaviour.

The strongest results on CPSC 2018 are obtained by the SMRD family rather than DBPD. For EfficientNetV2-S, standard SMRD with $\tau=0.3$ achieves a macro-F1 of 0.7166 at 23.76 effective epochs, while ERTS-SMRD improves this to 0.7188 and reduces the effective epoch count to 19.43. The same pattern holds for ResNet-18, where ERTS-SMRD improves macro-F1 from 0.7052 to 0.7088 while reducing EE from 23.76 to 19.39. MobileNetV2 follows the same trend, with ERTS-SMRD improving macro-F1 from 0.6833 to 0.6868 and reducing EE from 23.07 to 18.82. These results show that, on CPSC 2018, ERTS improves the best-performing PDD family across all three backbones.

Compared with PTB-XL, the absolute macro-F1 gains on CPSC 2018 are smaller, but they are highly consistent. This suggests that the dataset may contain fewer low-quality or weakly informative uncertain samples, leaving less room for explanation filtering to produce large performance gains. Nevertheless, ERTS consistently improves the efficiency--performance trade-off by achieving higher macro-F1 with fewer effective epochs. This is particularly important because it shows that the method is not only useful in noisier or more ambiguous settings, but also remains beneficial when the underlying PDD baseline is already strong.

The DBPD results provide additional insight into the effect of the focus threshold. For all three backbones, ERTS-DBPD with $\tau=0.7$ and $\phi=0.9$ gives the strongest DBPD-based macro-F1, improving over plain DBPD while keeping the number of effective epochs slightly lower. For example, EfficientNetV2-S improves from 0.6813 to 0.6861, ResNet-18 improves from 0.6708 to 0.6764, and MobileNetV2 improves from 0.6500 to 0.6554. As in the PTB-XL experiments, the more aggressive setting $\phi=0.5$ reduces computation more strongly but does not provide the best predictive performance, indicating that excessive filtering can remove useful hard examples.

The SRD results further support the role of explanation quality as a general reliability signal. Although SRD uses random sample reduction rather than confidence-based selection, adding ERTS generally improves macro-F1 while reducing the number of samples used for optimisation. This indicates that the Grad-CAM focus score is not merely duplicating the confidence signal used by DBPD. Instead, it provides an additional criterion for identifying samples whose uncertainty is less likely to be supported by coherent explanatory evidence.

Overall, the CPSC 2018 results reinforce the main conclusion of the paper. ERTS does not depend on a single dropout strategy or backbone architecture. On this dataset, its strongest effect appears when paired with SMRD, where it consistently improves macro-F1 and reduces effective epochs across all three backbones. At the same time, the DBPD and SRD results show that explanation-guided filtering also provides useful gains across other selection mechanisms. This supports the view that explanation quality can serve as a broadly applicable training-time reliability signal for efficient ECG classification.

\subsection{Georgia 2020 Results}
\begin{table*}[!t]
\centering
\caption{
Detailed Georgia 2020 results across EfficientNetV2-S, ResNet-18, and MobileNetV2. The table reports macro-F1, test accuracy, total samples used for backpropagation, Effective Epochs (EE), and the number of samples removed by the Grad-CAM explanation filter where applicable. We include the full-data baseline, standard PDD variants, and their corresponding ERTS-enhanced variants. Sample counts are abbreviated using K for thousands. Missing or unavailable entries are denoted by ``--''. Bold values indicate the best macro-F1 within each backbone.
}
\label{tab:georgia_all_backbones_full}
\resizebox{\textwidth}{!}{%
\begin{tabular}{l
ccccc
ccccc
ccccc}
\toprule
\multirow{2}{*}{Model}
& \multicolumn{5}{c}{EfficientNetV2-S}
& \multicolumn{5}{c}{ResNet-18}
& \multicolumn{5}{c}{MobileNetV2} \\
\cmidrule(lr){2-6}
\cmidrule(lr){7-11}
\cmidrule(lr){12-16}
& F1 & Acc. & Samples & EE & CAM
& F1 & Acc. & Samples & EE & CAM
& F1 & Acc. & Samples & EE & CAM \\
\midrule

Baseline
& 0.6375 & 74.69 & 828K & 100.0 & --
& 0.6248 & 73.95 & 828K & 100.0 & --
& 0.6054 & 72.32 & 828K & 100.0 & -- \\

\midrule

DBPD ($\tau=0.3$)
& 0.5790 & 63.12 & 177K & 21.39 & --
& 0.5665 & 62.18 & 177K & 21.39 & --
& 0.5490 & 60.79 & 172K & 20.75 & -- \\

ERTS-DBPD ($\tau=0.3,\phi=0.7$)
& 0.5864 & 63.65 & 138K & 16.71 & 39K
& 0.5738 & 62.79 & 138K & 16.73 & 39K
& 0.5561 & 61.39 & 134K & 16.23 & 38K \\

DBPD ($\tau=0.7$)
& 0.6657 & 75.56 & 334K & 40.32 & --
& 0.6521 & 74.83 & 334K & 40.32 & --
& 0.6319 & 73.18 & 324K & 39.11 & -- \\

ERTS-DBPD ($\tau=0.7,\phi=0.5$)
& 0.6578 & 74.92 & 245K & 29.58 & 89K
& 0.6436 & 74.21 & 245K & 29.59 & 89K
& 0.6236 & 72.57 & 238K & 28.70 & 86K \\

ERTS-DBPD ($\tau=0.7,\phi=0.7$)
& 0.6698 & 75.81 & 281K & 33.97 & 53K
& \textbf{0.6569} & 75.14 & 281K & 33.99 & 52K
& \textbf{0.6365} & 73.48 & 273K & 32.98 & 51K \\

ERTS-DBPD ($\tau=0.7,\phi=0.9$)
& \textbf{0.6736} & 76.02 & 316K & 38.24 & 17K
& -- & -- & -- & -- & --
& -- & -- & -- & -- & -- \\

\midrule

SMRD ($\tau=0.3$)
& 0.6277 & 74.40 & 196K & 23.73 & --
& 0.6150 & 73.62 & 196K & 23.73 & --
& 0.5959 & 72.00 & 191K & 23.03 & -- \\

ERTS-SMRD ($\tau=0.3,\phi=0.7$)
& 0.6339 & 74.73 & 152K & 18.36 & 45K
& 0.6216 & 74.01 & 152K & 18.36 & 45K
& 0.6023 & 72.38 & 147K & 17.81 & 43K \\

SMRD ($\tau=0.7$)
& 0.6548 & 76.62 & 343K & 41.47 & --
& 0.6418 & \textbf{75.89} & 343K & 41.47 & --
& 0.6219 & 74.21 & 333K & 40.23 & -- \\

ERTS-SMRD ($\tau=0.7,\phi=0.7$)
& 0.6607 & \textbf{76.84} & 267K & 32.23 & 76K
& 0.6487 & 76.20 & 267K & 32.24 & 76K
& 0.6286 & \textbf{74.52} & 259K & 31.27 & 74K \\

\midrule

SRD ($\gamma=0.95$)
& 0.6136 & 72.68 & 164K & 19.81 & --
& 0.6018 & 72.48 & 164K & 19.81 & --
& 0.5831 & 70.88 & 159K & 19.21 & -- \\

ERTS-SRD ($\gamma=0.95,\phi=0.7$)
& 0.6185 & 73.01 & 127K & 15.40 & 37K
& 0.6072 & 72.80 & 127K & 15.40 & 37K
& 0.5883 & 71.19 & 124K & 14.94 & 35K \\

SRD ($\gamma=0.98$)
& 0.6258 & 73.89 & 326K & 39.42 & --
& 0.6114 & 73.09 & 326K & 39.42 & --
& 0.5924 & 71.48 & 316K & 38.24 & -- \\

ERTS-SRD ($\gamma=0.98,\phi=0.7$)
& 0.6296 & 74.18 & 254K & 30.67 & 72K
& 0.6021 & 73.32 & 254K & 30.67 & 72K
& 0.5834 & 71.71 & 246K & 29.75 & 70K \\

SRD ($\gamma=0.99$)
& 0.6207 & 73.24 & 816K & 98.56 & --
& 0.6089 & 72.61 & 816K & 98.56 & --
& 0.5900 & 71.01 & 791K & 95.60 & -- \\

ERTS-SRD ($\gamma=0.99,\phi=0.7$)
& 0.6241 & 73.42 & 635K & 76.72 & 181K
& 0.6231 & 72.31 & 635K & 76.71 & 181K
& 0.6038 & 70.72 & 615K & 74.44 & 176K \\

\bottomrule
\end{tabular}%
}
\end{table*}

Table~\ref{tab:georgia_all_backbones_full} reports the detailed Georgia 2020 results across EfficientNetV2-S, ResNet-18, and MobileNetV2. This dataset provides an important additional test case because it is larger than CPSC 2018 and exhibits a different distribution of ECG labels and recording characteristics. As in the previous analyses, we compare the full-data baseline against DBPD, SMRD, and SRD, together with the corresponding ERTS-enhanced variants. This allows us to evaluate whether explanation-guided filtering remains effective when the underlying data distribution and optimal dropout behaviour differ from PTB-XL and CPSC 2018.

On Georgia 2020, the strongest macro-F1 results are obtained by the ERTS-DBPD family. For EfficientNetV2-S, plain DBPD with $\tau=0.7$ achieves a macro-F1 of 0.6657 at 40.32 effective epochs, while ERTS-DBPD with $\phi=0.9$ improves macro-F1 to 0.6736 and reduces EE to 38.24. This represents the best overall result for this backbone. For ResNet-18, ERTS-DBPD with $\phi=0.7$ improves over plain DBPD, increasing macro-F1 from 0.6521 to 0.6569 while reducing EE from 40.32 to 33.99. The same pattern is observed for MobileNetV2, where ERTS-DBPD improves macro-F1 from 0.6319 to 0.6365 and reduces EE from 39.11 to 32.98. These results show that ERTS improves the best DBPD-based trade-off across all three model architectures.

The threshold behaviour on Georgia 2020 differs slightly from PTB-XL. While $\phi=0.9$ gives the best EfficientNetV2-S result, $\phi=0.7$ provides the best available ERTS-DBPD performance for ResNet-18 and MobileNetV2. This suggests that the optimal strength of explanation filtering can depend on both dataset properties and model capacity. The larger EfficientNetV2-S backbone appears to benefit from a more selective filter that removes only the most diffuse explanations, whereas the smaller architectures obtain stronger results with a moderately broader filtering criterion. Nevertheless, the overall trend remains consistent: ERTS improves macro-F1 over plain DBPD while reducing the number of effective epochs.

The SMRD results provide further evidence that explanation-guided filtering improves efficiency even when the predictive gain is more modest. For EfficientNetV2-S, ERTS-SMRD with $\tau=0.7$ improves macro-F1 from 0.6548 to 0.6607 and reduces EE from 41.47 to 32.23. For ResNet-18, ERTS-SMRD improves macro-F1 from 0.6418 to 0.6487 with a similar reduction in EE, and for MobileNetV2 it improves macro-F1 from 0.6219 to 0.6286 while reducing EE from 40.23 to 31.27. These results indicate that explanation quality can refine not only confidence-driven sample selection, but also schedule-matched random selection.

The SRD family shows a more nuanced pattern. For $\gamma=0.95$ and $\gamma=0.99$, ERTS-SRD improves macro-F1 across all three backbones while reducing the number of samples used for optimisation. However, for $\gamma=0.98$, ERTS-SRD improves EfficientNetV2-S but reduces macro-F1 for ResNet-18 and MobileNetV2, despite improving accuracy slightly. This suggests that explanation filtering is not uniformly beneficial under every random dropout schedule, particularly for smaller models where random retention may already remove useful variation. Still, the consistent reductions in total samples and EE show that the filter remains effective at reducing optimisation cost.

Overall, the Georgia 2020 results reinforce the main conclusion drawn from the previous datasets. ERTS consistently improves the efficiency--performance trade-off when paired with the strongest PDD strategy for each backbone, especially DBPD. The gains are achieved without increasing model complexity or changing the architecture, indicating that explanation quality provides a useful training-time reliability signal. At the same time, the differences between $\phi=0.7$ and $\phi=0.9$ show that the filtering threshold should be treated as a dataset- and model-dependent hyperparameter rather than a fixed universal choice.

\subsection{Effect of Explainability Filtering}

The results across PTB-XL, CPSC 2018, and Georgia 2020 show that the benefit of ERTS depends strongly on how the explanation filter is applied. In particular, the focus threshold $\phi$ controls how selectively the method removes uncertain samples after the initial PDD stage. A lower threshold applies a more aggressive filter and removes a larger portion of the candidate set, while a higher threshold acts more conservatively by filtering only the samples with the weakest explanatory structure. Across the three datasets, the strongest results are generally obtained when the explanation filter is used as a selective reliability mechanism rather than as a broad pruning strategy.

This trend is clearest in the DBPD experiments. On PTB-XL, ERTS-DBPD with $\phi=0.5$ consistently reduces the effective epoch count, but it also substantially degrades macro-F1 across all three backbones. For EfficientNetV2-S, macro-F1 falls from 0.6488 with plain DBPD to 0.5915 with ERTS-DBPD at $\phi=0.5$. Similar drops are observed for ResNet-18 and MobileNetV2. This indicates that overly aggressive filtering removes not only unreliable low-focus samples, but also hard examples that remain informative for learning. In contrast, $\phi=0.9$ provides the best DBPD-based trade-off on PTB-XL, improving macro-F1 across EfficientNetV2-S, ResNet-18, and MobileNetV2 while reducing EE relative to plain DBPD.

The CPSC 2018 results show a similar but more stable pattern. For the DBPD family, increasing the focus threshold from $\phi=0.5$ to $\phi=0.9$ progressively improves macro-F1 across all three backbones. For example, with EfficientNetV2-S, ERTS-DBPD improves from 0.6752 at $\phi=0.5$ to 0.6861 at $\phi=0.9$. ResNet-18 and MobileNetV2 follow the same trend. However, the strongest overall CPSC 2018 results are obtained by ERTS-SMRD with $\tau=0.3$ and $\phi=0.7$, which improves macro-F1 over plain SMRD while reducing EE by approximately 18\% across all three backbones. This suggests that when the base sampling strategy already provides a strong training subset, a moderate explanation filter is sufficient to remove unreliable samples without discarding useful variation.

On Georgia 2020, the optimal threshold is more dependent on model capacity. EfficientNetV2-S benefits most from the more selective $\phi=0.9$ setting, where ERTS-DBPD improves macro-F1 from 0.6657 to 0.6736 while reducing EE from 40.32 to 38.24. For ResNet-18 and MobileNetV2, the best available ERTS-DBPD configuration uses $\phi=0.7$, improving macro-F1 from 0.6521 to 0.6569 and from 0.6319 to 0.6365, respectively, while reducing EE by more than six effective epochs in each case. This suggests that the ideal amount of filtering depends not only on dataset properties, but also on the capacity of the model. Larger models may tolerate a more selective explanation criterion, whereas smaller models may benefit from a slightly broader retention regime.

The results also clarify the role of ERTS beyond confidence-based selection. ERTS improves or reduces the cost of SMRD and SRD variants in many settings, even though these methods are not driven by the same confidence criterion as DBPD. This indicates that the Grad-CAM focus score is not merely duplicating model confidence. Instead, it provides an additional signal about whether the model's prediction is supported by coherent explanatory evidence. The SRD results are especially informative because the underlying selection strategy is random. Improvements in several ERTS-SRD configurations suggest that explanation quality can act as a general reliability signal, although the Georgia 2020 results also show that the benefit can depend on the random dropout schedule and model capacity.

Overall, these findings indicate that explanation quality is most effective when used as a reliability filter rather than as a strict pruning mechanism. The aim of ERTS is not to minimise the number of training samples as aggressively as possible, but to identify which uncertain samples are likely to be useful for further optimisation. Samples with diffuse or weak Grad-CAM evidence are less reliable and can often be removed without harming performance, while uncertain samples with coherent focus should be retained. Across datasets and backbones, this selective use of explanation filtering improves the efficiency--performance trade-off, with the best ERTS configurations reducing effective epochs by roughly 5\%--18\% relative to their strongest plain PDD counterparts while maintaining or improving macro-F1.

\subsection{Training Efficiency Analysis}

While Effective Epochs provide a normalised measure of training efficiency, they do not directly show the absolute volume of data used during optimisation. Figure~\ref{fig:samples_seen} complements the EE analysis by reporting the total number of samples that contribute to backpropagation for each method. This is important because two methods may appear close in terms of predictive performance while requiring substantially different numbers of gradient-update samples. In practical clinical model development, this difference directly affects training time, energy consumption, and the feasibility of repeated retraining across datasets or institutions.

The figure shows that the full-data baseline processes the largest number of samples, as expected, because every training example is used for gradient updates at every epoch. Standard PDD variants reduce this cost by progressively excluding samples that are considered learned or no longer necessary for optimisation. However, ERTS further reduces the training workload by applying an additional explanation-based reliability filter to the candidate samples selected by PDD. As a result, ERTS variants generally process fewer samples than their corresponding non-explainability counterparts.

\inlinefigure[0.95\linewidth]
{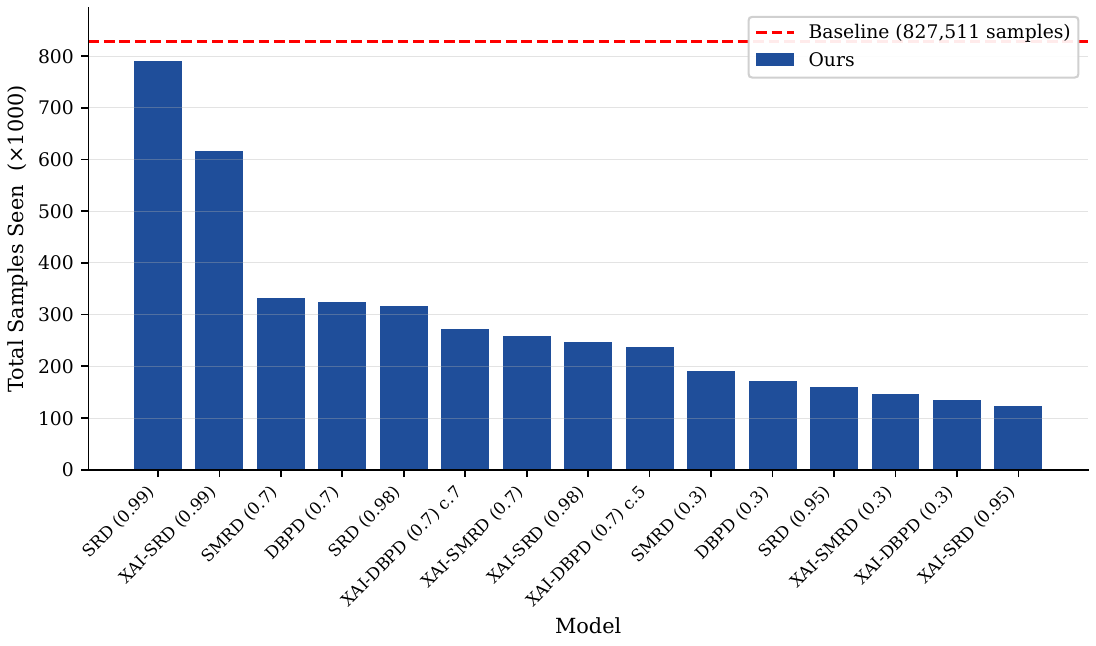}
{
Absolute number of training samples used for gradient updates across the baseline, standard PDD variants, and ERTS-enhanced variants. The dashed reference line denotes the full-data baseline. Lower bars indicate fewer samples contributing to backpropagation and therefore lower optimisation cost. ERTS variants generally reduce the number of samples processed relative to their corresponding PDD counterparts.
}
{fig:samples_seen}

This reduction is particularly clear for the DBPD and SMRD families. ERTS-DBPD and ERTS-SMRD require substantially fewer gradient-update samples than the baseline, while the corresponding results in Tables~\ref{tab:ptbxl_all_backbones_full}, \ref{tab:cpsc_all_backbones_full}, and \ref{tab:georgia_all_backbones_full} show that these reductions are achieved without sacrificing predictive performance in the best-performing configurations. In several cases, ERTS improves macro-F1 while simultaneously reducing the total number of samples used for optimisation. This confirms that ERTS is not merely shifting computation from one part of training to another, but is reducing redundant gradient updates in a way that improves the efficiency--performance trade-off.

The sample-count analysis also helps explain why ERTS is practically useful. Effective epochs are useful for comparing methods independently of hardware, but total sample counts give a more concrete view of training workload. By filtering samples whose predictions are not supported by focused explanations, ERTS reduces the amount of data passed through backpropagation. This is especially relevant for clinical ECG learning, where models may need to be retrained under different label definitions, population distributions, or deployment constraints. Overall, Fig.~\ref{fig:samples_seen} shows that the improvements reported throughout the experimental section correspond to meaningful reductions in optimisation cost, not only marginal changes in an abstract efficiency metric.

\subsection{Two-Stage Filtering Behaviour}

Figure~\ref{fig:confidence_focus} provides a direct visualisation of the two-stage filtering mechanism used by ERTS. The horizontal axis represents the model confidence for the true class, while the vertical axis represents the Grad-CAM focus score. In the first stage, the PDD confidence threshold identifies candidate samples that remain uncertain and may therefore still be useful for learning. In the second stage, ERTS evaluates these candidates using explanation quality, retaining samples whose predictions are supported by sufficiently focused Grad-CAM evidence and filtering those with weak or diffuse attention.

\inlinefigure[0.95\linewidth]
{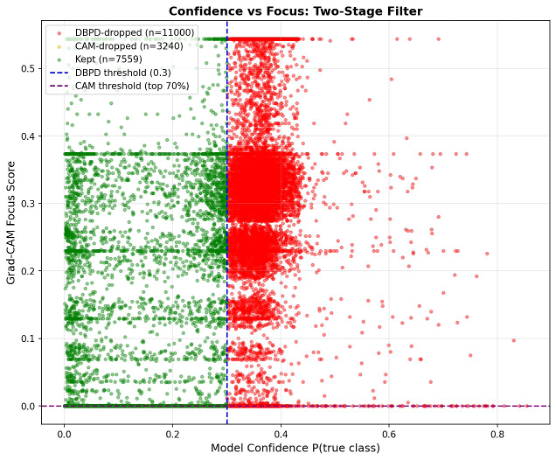}
{
Two-stage filtering behaviour of ERTS in the confidence--focus space. The vertical dashed line denotes the PDD confidence threshold, while the horizontal dashed line denotes the Grad-CAM focus threshold. Green points indicate samples retained for training, yellow points indicate samples removed by the Grad-CAM explanation filter, and red points indicate samples removed by confidence-based DBPD filtering.
}
{fig:confidence_focus}

This plot helps clarify why ERTS differs from confidence-only data selection. Confidence alone treats all samples below the threshold as similarly useful, even though their explanatory structure may differ substantially. By incorporating the focus score, ERTS separates low-confidence samples into retained and CAM-filtered groups. Retained samples occupy the uncertain but higher-focus region, suggesting that they remain difficult while still being supported by coherent evidence. In contrast, CAM-filtered samples are concentrated near the low-focus region, indicating that their uncertainty is less reliably grounded in salient ECG structure.

The red DBPD-filtered samples provide an additional reference point. These samples are removed by the confidence-based stage and therefore do not proceed to the explanation-filtering step. The separation between retained, CAM-filtered, and DBPD-filtered samples supports the central design of ERTS: rather than simply reducing the number of training examples, the method uses explanation quality to decide which uncertain examples should continue to influence optimisation.

\subsection{Class-Level Analysis of Sample Selection}

\begin{figure*}[!t]
\centering
\includegraphics[width=0.9\textwidth]{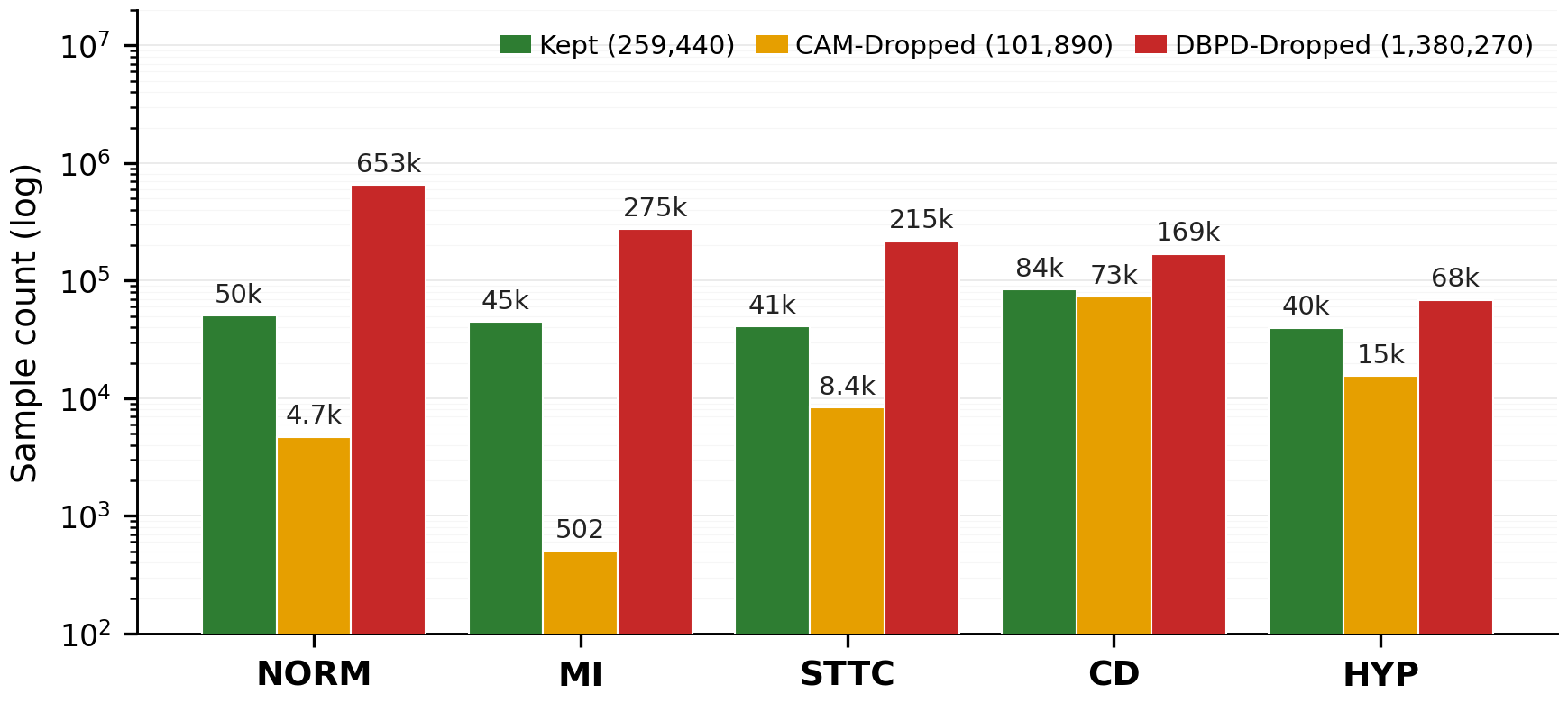}
\caption{
Class distribution of samples retained by ERTS, removed by the Grad-CAM filter, and removed by confidence-based DBPD filtering. Results are shown for the PTB-XL EfficientNetV2-S model using $\tau=0.3$ and $\phi=0.7$.
}
\label{fig:class_distribution}
\end{figure*}

To better understand which samples are being removed, Fig.~\ref{fig:class_distribution} shows the class distribution of retained samples, CAM-filtered samples, and confidence-filtered samples. Several interesting patterns emerge. First, the Grad-CAM filter removes a large number of NORM recordings, accounting for 3,490 of the 6,446 CAM-filtered samples (54.1\%). This suggests that normal ECGs frequently produce diffuse explanations, likely because they lack a focal pathological anchor. In contrast, confidence-based filtering removes very few samples overall, indicating that many difficult examples remain available for further analysis.

More importantly, diagnostically relevant classes such as MI, STTC, and CD are largely retained by ERTS. This behaviour supports the central hypothesis of the paper: explanation quality acts as a reliability signal that preferentially removes samples with weak explanatory structure while preserving clinically meaningful uncertainty.

\begin{figure*}[!t]
\centering
\includegraphics[width=0.9\textwidth]{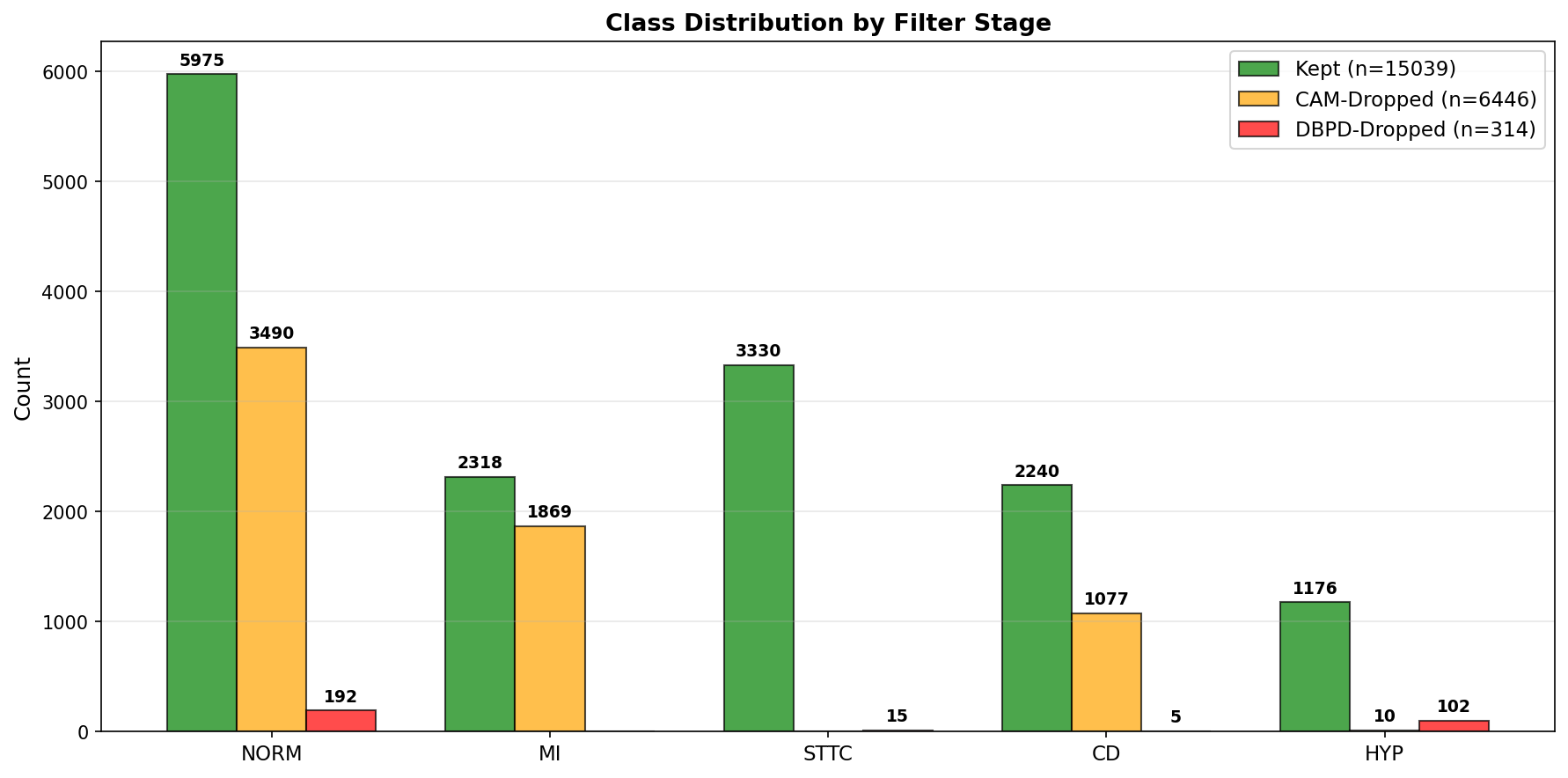}
\caption{
Log-scale view of cumulative sample counts retained by ERTS, removed by the Grad-CAM filter, and removed by confidence-based DBPD filtering throughout training. The logarithmic scale highlights large differences in filtering behaviour across classes.
}
\label{fig:class_distribution_log}
\end{figure*}

Figure~\ref{fig:class_distribution_log} provides a cumulative view of sample-selection behaviour across the entire training process. A striking observation is that confidence-based filtering removes substantially larger numbers of samples from MI and STTC classes than the explainability filter. For example, over 650k NORM samples and more than 275k MI samples are removed by DBPD, whereas ERTS removes only a small fraction of these examples through explanation filtering.

This suggests that confidence-based selection tends to discard classes that are learned early, regardless of their clinical importance. In contrast, ERTS acts as a targeted reliability filter, removing samples with diffuse explanations while preserving diagnostically meaningful examples for continued optimisation. These findings provide mechanistic evidence for why explanation-guided selection leads to improved performance despite using fewer effective training samples.

\subsection{Qualitative Analysis of Retained and Filtered Samples}

To further inspect the behaviour of the explanation filter, we qualitatively compare representative samples retained by ERTS with samples removed by the Grad-CAM focus criterion. These examples complement the quantitative filtering results by showing how the focus score reflects the structure of model attention. In retained samples, the Grad-CAM maps tend to concentrate around temporally localised waveform regions or diagnostically meaningful lead patterns. In contrast, filtered samples often exhibit diffuse or unstable attention spread across the recording, suggesting weaker explanatory support for continued optimisation.

\begin{figure*}[!t]
\centering

\begin{subfigure}[!t]{0.48\textwidth}
\centering
\includegraphics[width=\linewidth]{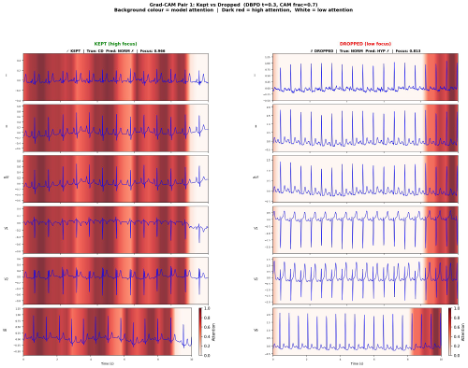}
\caption{
Kept CD sample with focused QRS-locked attention across leads I, II, and aVF, compared with a dropped NORM sample showing diffuse attention across the full recording.
}
\label{fig:qual_case_cd_norm}
\end{subfigure}
\hfill
\begin{subfigure}[!t]{0.48\textwidth}
\centering
\includegraphics[width=\linewidth]{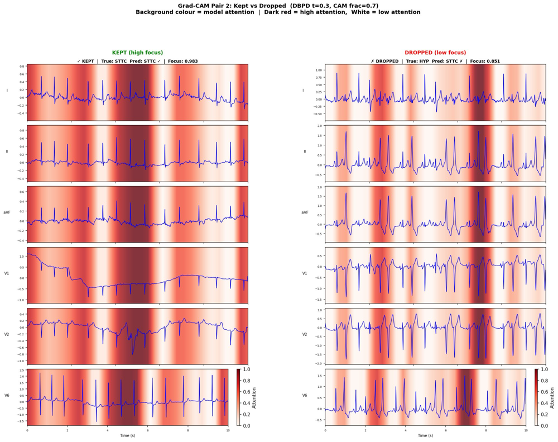}
\caption{
Kept STTC sample with tight attention near the J-point, compared with a dropped HYP sample showing broader attention across precordial leads.
}
\label{fig:qual_case_sttc_hyp}
\end{subfigure}

\vspace{0.8em}

\begin{subfigure}[!t]{0.48\textwidth}
\centering
\includegraphics[width=\linewidth]{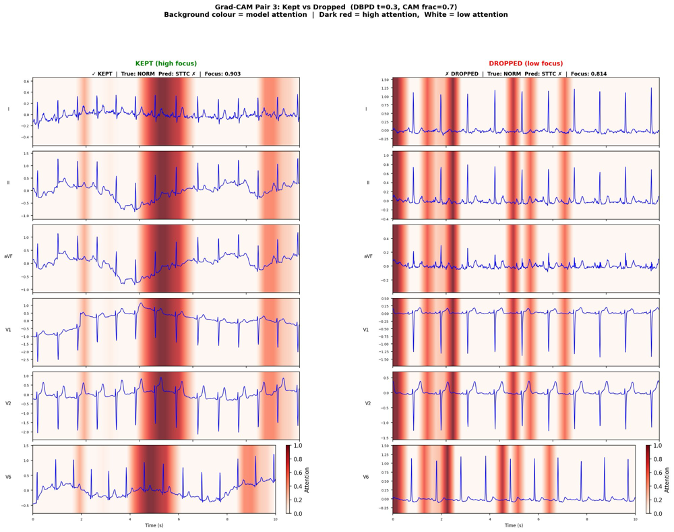}
\caption{
Comparison of two NORM samples. The retained sample localises attention to a compact temporal window, whereas the dropped sample shows oscillatory and diffuse attention.
}
\label{fig:qual_case_norm_norm}
\end{subfigure}
\hfill
\begin{subfigure}[!t]{0.48\textwidth}
\centering
\includegraphics[width=\linewidth]{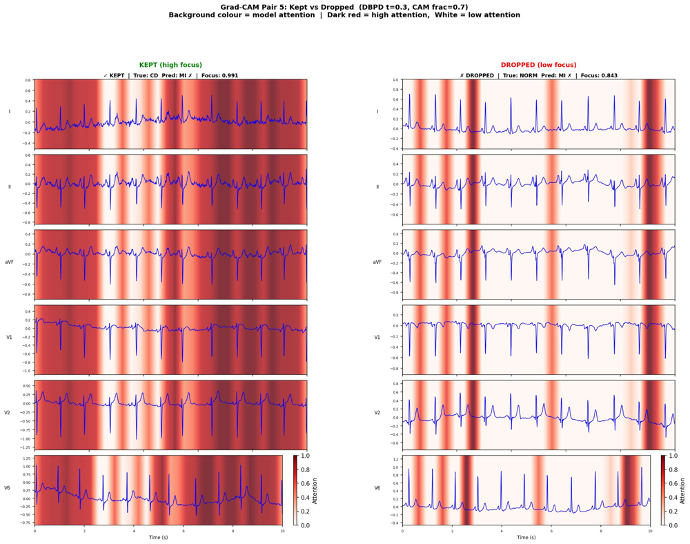}
\caption{
Kept CD sample with strong QRS-locked attention in leads I and V2, compared with a dropped NORM sample with diffuse attention and no clear temporal peak.
}
\label{fig:qual_case_cd_norm_2}
\end{subfigure}

\caption{
Qualitative examples of samples retained and filtered by ERTS. Each panel contrasts a retained sample with a dropped sample using Grad-CAM attention maps and focus scores. Retained samples generally show more coherent and localised attention around ECG morphology, while dropped samples show more diffuse activation patterns. These examples illustrate how ERTS uses explanation quality to distinguish informative uncertainty from less reliable training signal.
}
\label{fig:qualitative_examples}
\end{figure*}

Across the four examples in Fig.~\ref{fig:qualitative_examples}, retained samples show more coherent explanatory structure than filtered samples. For conduction disturbance and ST-T change cases, attention is concentrated around waveform regions that are plausibly relevant to the predicted class. This supports the interpretation that ERTS preserves difficult samples when uncertainty is accompanied by meaningful evidence. In contrast, several filtered samples exhibit attention distributed broadly across time or across multiple leads without a clear temporal focus. Such patterns suggest that the sample may provide a weaker or less stable gradient signal, even if the model remains uncertain.

These qualitative examples also highlight why ERTS should not be interpreted as simply removing normal samples or minority classes. The comparison between two NORM examples shows that samples from the same class can be treated differently depending on the quality of their explanations. A retained normal recording can still display a compact and interpretable attention pattern, while another normal recording may be filtered because its Grad-CAM map is diffuse. This reinforces the central premise of ERTS: the filtering decision is based not only on class label or confidence, but on whether the model's uncertainty is supported by focused explanatory evidence.

\FloatBarrier

\section{Discussion}

The results show that explanation quality can be used as a practical training-time signal for efficient ECG classification. Across three datasets and three backbone architectures, ERTS improves the best-performing PDD configuration while reducing effective epochs. This suggests that the method is not tied to a specific dataset, architecture, or dropout strategy, but provides a general way to refine progressive data selection.

A key observation is that model confidence alone is not sufficient to determine whether a difficult sample is useful. In ECG data, low confidence may correspond to clinically meaningful but under-learned patterns, or it may arise from noise, ambiguity, artefacts, or weak label quality. ERTS addresses this distinction by adding an explanation-based reliability check. Samples are retained only when their uncertainty is supported by focused Grad-CAM evidence, while low-focus samples are filtered out.

The threshold analysis shows that the strength of this filtering is important. Aggressive filtering can reduce computation, but it may also remove informative hard examples and degrade macro-F1. This is most visible when $\phi=0.5$, where effective epochs decrease but performance often drops. In contrast, moderate or selective filtering provides a better balance between performance and efficiency. Across the three datasets evaluated, $\phi=0.7$ provides a reliable default that avoids aggressive pruning while still filtering low-quality samples, while $\phi=0.9$ appears useful when the backbone has higher representational capacity. This indicates that ERTS should be viewed as a reliability filter rather than a strict pruning method.

ERTS also improves or reduces the cost of SMRD and SRD variants in several settings, indicating that the Grad-CAM focus score captures information beyond model confidence. This is important because SRD does not rely on confidence-based selection. The improvement of ERTS-SRD in multiple cases suggests that explanation quality can act as a broader signal of sample reliability, not merely as a refinement of DBPD.

From a practical perspective, ERTS can be integrated into existing training pipelines with minimal architectural change. The method does not require modifying the backbone network or adding an auxiliary prediction head. Instead, it only changes the training loop by applying Grad-CAM to the candidate samples already selected by the PDD stage and then deciding whether those samples should contribute to backpropagation. This makes the method suitable for repeated retraining scenarios, where models may need to be updated across institutions, patient populations, or label definitions. The reduction in gradient-update samples also connects directly to the computational motivation of this work: fewer backpropagated samples imply lower optimisation cost, shorter retraining cycles, and potentially reduced energy use.

The main limitation of this work is that ERTS currently relies on Grad-CAM and a simple focus-score criterion. Grad-CAM was chosen because it is computed per class without requiring input perturbation, making it compatible with a per-epoch filtering loop. Perturbation-based methods such as LIME would be prohibitively expensive during training, while more fine-grained attribution methods such as Integrated Gradients would add additional cost even though the proposed focus score only requires a relative concentration measure rather than pixel-precise attribution. Grad-CAM's spatial aggregation is therefore sufficient for the binary keep/drop decision used in ERTS. Nevertheless, explanation quality may not always be equivalent to spatial or temporal concentration, since some ECG abnormalities may involve distributed patterns across time or leads. Future work should therefore explore Grad-CAM++, Integrated Gradients, adaptive thresholding, and clinically informed explanation-quality measures. Further validation with expert review of retained and filtered samples would also help clarify whether low-focus examples correspond to artefacts, ambiguous labels, or genuinely difficult cases.

Overall, ERTS extends the role of explainability from post hoc interpretation to training-time optimisation. Rather than only asking whether a sample is difficult, the method asks whether it is difficult for a meaningful reason. This provides a simple but effective route toward more efficient and reliable clinical time-series learning.

\section{Declaration of generative AI and AI-assisted technologies in the manuscript preparation process}

During the preparation of this work, the author(s) used ChatGPT for fixing grammar and writing style. The author(s) reviewed and edited the output as needed and take full responsibility for the content of the published article.

\section{Conclusion}

In this work, we introduced ERTS, an explainability-guided data selection framework for efficient clinical time-series learning. The central idea is to use explanation quality as a training-time reliability signal, enabling the model to distinguish between informative and unreliable uncertainty. By integrating this signal with Progressive Data Dropout, ERTS refines the set of samples used for gradient updates, prioritising those that exhibit both low confidence and meaningful attention patterns.

Across multiple ECG datasets and backbone architectures, ERTS consistently improves the trade-off between predictive performance and training efficiency. The method achieves higher macro-F1 while reducing the number of effective epochs required for optimisation. Notably, improvements are observed not only for confidence-based selection strategies but also when combined with random sampling, suggesting that explanation quality acts as a general-purpose signal for filtering unreliable training data.

These findings highlight a broader perspective on explainability in clinical machine learning. Rather than serving solely as a post hoc interpretability tool, explanation methods can play an active role during optimisation by guiding how models learn from data. This is particularly important in clinical settings, where data quality, label ambiguity, and resource constraints necessitate both efficient and trustworthy learning strategies.

Future work will explore alternative explanation methods, adaptive thresholding strategies, and extensions to other modalities such as multimodal clinical data and long-form physiological signals. More broadly, we believe that integrating interpretability into the training loop offers a promising direction for developing data-efficient and reliable machine learning systems in healthcare.

\bibliographystyle{cas-model2-names}

\bibliography{main}

\iffalse
\bio{}
Author biography without author photo.
Author biography. Author biography. Author biography.
Author biography. Author biography. Author biography.
Author biography. Author biography. Author biography.
Author biography. Author biography. Author biography.
Author biography. Author biography. Author biography.
Author biography. Author biography. Author biography.
Author biography. Author biography. Author biography.
Author biography. Author biography. Author biography.
Author biography. Author biography. Author biography.
\endbio

\bio{figs/cas-pic1}
Author biography with author photo.
Author biography. Author biography. Author biography.
Author biography. Author biography. Author biography.
Author biography. Author biography. Author biography.
Author biography. Author biography. Author biography.
Author biography. Author biography. Author biography.
Author biography. Author biography. Author biography.
Author biography. Author biography. Author biography.
Author biography. Author biography. Author biography.
Author biography. Author biography. Author biography.
\endbio

\bio{figs/cas-pic1}
Author biography with author photo.
Author biography. Author biography. Author biography.
Author biography. Author biography. Author biography.
Author biography. Author biography. Author biography.
Author biography. Author biography. Author biography.
\endbio
\fi

\end{document}